# Sarcasm Detection in Twitter - Performance Impact while using Data Augmentation: Word Embeddings

Alif Tri Handoyo
Computer Science Department
BINUS Graduate Program,
Master of Computer Science,
*Bina Nusantara University*,
Jakarta 11480, Indonesia
alif.handoyo@binus.edu

Hidayaturrahman
Computer Science Department
School of Computer Science,
*Bina Nusantara University,*
Jakarta 11480, Indonesia
hidayaturrahman@binus.ac.id

Derwin Suhartono
Computer Science Department
School of Computer Science,
*Bina Nusantara University,*
Jakarta 11480, Indonesia
dsuhartono@binus.edu

*Abstract*— Sarcasm is the use of words usually used to either mock or annoy someone, or for humorous purposes. Sarcasm is largely used in social networks and microblogging websites, where people mock or censure in a way that makes it difficult even for humans to tell if what is said is what is meant. Failure to identify sarcastic utterances in Natural Language Processing applications such as sentiment analysis and opinion mining will confuse classification algorithms and generate false results. Several studies on sarcasm detection have utilized different learning algorithms. However, most of these learning models have always focused on the contents of expression only, leaving the contextual information in isolation. As a result, they failed to capture the contextual information in the sarcastic expression. Moreover, some datasets used in several studies have an unbalanced dataset which impacting the model result. In this paper, we propose a contextual model for sarcasm identification in twitter using RoBERTa, and augmenting the dataset by applying Global Vector representation (GloVe) for the construction of word embedding and context learning to generate more data and balancing the dataset. The effectiveness of this technique is tested with various datasets and data augmentation settings. In particular, we achieve performance gain by 3.2% in the iSarcasm dataset when using data augmentation to increase 20% of data labeled as sarcastic, resulting F-score of 40.4% compared to 37.2% without data augmentation.

I. INTRODUCTION

Twitter has become one of the biggest web destinations for users to express their opinions and thoughts. Many companies and organizations have been interested in these data to study the opinion of people towards political events [1], popular products [2], or movies [3].

However, due to the informal language used in Twitter and the limitation in terms of characters (i.e., 140 characters per tweet), understanding the opinions of users and performing such analysis is quite difficult. Furthermore, the presence of sarcasm makes the task even more challenging: sarcasm is when a person says something different from what he means.

Some people are more sarcastic than others, however, in general, sarcasm is very common, though, difficult to recognize [4]. Sarcasm is a form of irony that occurs when there is some discrepancy between the literal and intended meanings of an utterance. This discrepancy is used to express dissociation towards a previous proposition, often in the form of contempt or derogation [5]. However, in general, people employ sarcasm in their daily life not only to make jokes and be humorous but also to criticize or make remarks about ideas, persons, or events. Therefore, it tends to be widely used in social networks, in particular microblogging websites such as Twitter.

A challenge specific to sarcasm detection is the difficulty in acquiring ground-truth annotations. Human-annotated datasets usually contain only a few thousand texts, resulting in many small datasets. In comparison, automatic data collection using distant supervision signals like hashtags yielded substantially larger datasets [6]. Nevertheless, the automatic approach also led to label noise. For example, it is found nearly half of the tweets with sarcasm hashtags in one dataset are non-sarcastic. There is also a noise-free example of a dataset where the sarcasm tweets are labeled by their authors but having a huge unbalanced data resulting in a very low performance [5].

In this paper, we analyze the performance impact when using data augmentation in four different datasets, iSarcasm [5], Ghosh [7], Ptacek [8], and SemEval-18 [9]. Data augmentation is applied in the sarcastic labeled data to increase the size of data. GloVe [10] was used for word embeddings to generate more data by replacing similar words in a sentence like "good" to "happy" that express similar meanings. We use the more recent neural representation model RoBERTa [11] to predict tweets as sarcasm. The results will be compared based on the model without data augmentation and model with data augmentation to increase sarcastic labeled data by different sizes respectively 10%, 20%, and 30% increased data.

The remainder of this paper is structured as follows. Section II describes some of the related work. Section III describes in detail our experiment method. Section IV illustrates our experiments and results, and Section V concludes the result of this work.

II. RELATED WORK

*A. Sarcasm Detection*

Acquiring large and reliable datasets has been quite a challenge for sarcasm detection. Due to the cost of annotation, manually labeled datasets such as iSarcasm [5] and SemEval-18 [9] typically contain only a few thousand data and have an unbalanced dataset. Automatic crawling datasets such as Ghosh [7] and Ptacek [8] generate much more data but the result is considerably noisy. As a case study, after examining some automatic crawling datasets, it is found that nearly half of the tweets with sarcastic labels are non-sarcastic [5].

Sarcasm identification task has been studied by employing different methods, including lexicon-based, conventional machine learning, deep learning, or even a hybrid approach. Besides, several reviews on sarcasm detection have also been conducted. For instance, there was a study that performed SLR on sarcasm identification in Textual data [12]. The study was carried out by considering 'dataset collection, preprocessing techniques, feature engineering techniques (feature extraction, feature selection, and feature representation), classification algorithms, and performance measures.' The study revealed that content-based features are the most employed features for sarcasm classification. The study also revealed that the standard evaluation metrics such as precision, recall, accuracy, f-measure, and Area under the curve (AUC) are the most used parameters for evaluating classifiers' performance. Moreover, the study also revealed that when there is an imbalance in the class distribution of the dataset, the AUC performance measure is the right choice due to its robustness in resisting the skewness in the dataset. The review concluded by identifying recent challenges and proposing the open research direction to provide a solution to the sarcasm identification studies issue.

Various scholars have studied sarcasm identification tasks. There was a study that stated two methods, namely, the ''Incongruent words-only'' method and ''all-words: method'' in ''Expect the Unexpected: Harnessing Sentence Completion for Sarcasm Detection'' research by employing ''Sentence completion'' for sarcastic analysis [13]. For evaluation purposes, two sets of data were used, which include (i) Twitter data collected by Riloff, et al. [14], consisting 2278 tweets ('506 sarcastic, and 1772 non-sarcastic). (ii) Discussion forum data collected by [15], that contains manually labeled balanced tweets ('752 sarcastic and 752 non-sarcastic). However, 'WordNet similarities and word2vec' were employed to measure the similarities in the performance. Two-fold cross-validation was used for evaluation purposes. Thus, the overall predictive results attained an F-score of 54% by employing the Word2Vec similarity for the all-words method, but 80.24% of F-score was obtained with the WordNet incongruous words-only method. On the other hand, an 80.28% F-score is obtained using the WordNet similarity and incongruous words-only method with 2-fold cross-validation.

*B. RoBERTa*

The optimized model called Robustly Optimized BERT Approach (RoBERTa), used 10 times more data (160GB compared with the 16GB originally exploited) and is trained with far more epochs than the BERT model (500K vs 100K), using also 8-times larger batch sizes, and a byte-level BPE vocabulary instead of the character-level vocabulary that was previously utilized. Another significant modification was the dynamic masking technique instead of the single static mask used in BERT. In addition, RoBERTa model removes the next sentence prediction objective used in BERT, following advice by several other studies that question the NSP loss term [11].

RoBERTa [11] have been used in various sarcasm identification tasks and achieve a satisfactory result. A study performed a contextual embedding method with RoBERTa [11] and obtained an 82.44% F-score [16]. A study using RoBERTa [11] in sarcasm identification task with Context Separators method achieves 77.2% F-score [17]. A study in Transformers on sarcasm detection with context, obtained the highest performance results using RoBERTa [11] with a 77.22% F-score [27]. Another study in Transformer-based Context-aware Sarcasm Detection in Conversation Threads from Social-Media obtained a great result with 78.3% F-Score for Twitter and 74.4% F-Score for reddit datasets [28].

*C. Data Augmentation*

Data Augmentation in the field of Image Processing and Computer Vision is a well-known methodology to increase the dataset by introducing varied distributions and increase the performance of the model for a number of tasks. In general, it is believed that the more data a neural network gets trained on, the more effective it becomes. Augmentations are performed by using simple image transformations such as rotation, cropping, flipping, translation, and addition of Gaussian noise to the image. A study used data augmentation methods to increase the training data size for training a deep neural network on the ImageNet [19] dataset. The increase in training data samples showed reduced overfitting of the model and increased the model performance [18]. These techniques enable the model to learn additional patterns in the image and identify new positional aspects of objects in it.

On similar lines, data augmentation methods are explored in the field of text processing for improving the efficacy of models. A study was performed by replaced random words in a sentence with their respective synonyms, to generate augmented data and train a siamese recurrent network for sentence similarity tasks [20]. Another study used word embeddings of sentences to generate augmented data to increase data size and trained a multi-class classifier on tweet data. They found the nearest neighbor of a word vector by using cosine similarity and used that as a replacement for the original word. The word selection was done stochastically [21].

For information extraction, A study [22] applied data augmentation techniques on legal dataset [23]. A class-specific probability classifier was trained to identify a particular contract element type for each token in a sentence. They classified a token in a sentence based on the window of words/tokens surrounding them. They used word embeddings obtained by pre-training a word2vec model [24] on unlabeled contract data. Their work examined three data augmentation methods namely; interpolation, extrapolation, and random noise. The augmentation methods manipulated the word embeddings to obtain a new set of sentence representations. The study also highlighted that the interpolation method performed comparatively better than the other methods like extrapolation and random noise [22]. Another study explored various perturbation methods where they introduced random perturbations like gaussian noise or bernouli noise into the word embeddings in text-related classification tasks such as sentence classification, sentiment classification, and relation classification [25].

III. METHODOLOGY

Given a set of tweets, we aim to classify each one of them depending on whether it is sarcastic or not. We use data augmentation to increase 10%, 20%, and 30% sarcastic labeled data, respectively. Then from each tweet, we extract a set of features, refer to a training set, and use RoBERTa [11] to perform the classification. The features are extracted in a way that covers the different types of sarcasm we identified.

*A. Data*

We conduct sarcasm detection experiments using two automatic datasets and two manually annotated datasets. The two automatically-collected datasets include Ptacek [8] and Ghosh [7], which treat tweets having particular hashtags such as #sarcastic, #sarcasm, #not as sarcastic, and others as non-sarcastic.

The two manually annotated datasets include iSarcasm [5] and SemEval-18 [9]. iSarcasm [5] dataset contains tweets written by participants of an online survey and thus is an example of intended sarcasm detection while SemEval-18 [9] consists of both sarcastic and ironic tweets supervised by third-party annotators and thus is used for perceived sarcasm detection.

**TABLE 1. Datasets statistics, including the number of samples in each split and the proportion of sarcastic texts.**

| Dataset | Train | Val | Test | % Sarcasm |
|---|---|---|---|---|
| iSarcasm | 3,116 | 347 | 887 | 17.62% |
| Ghosh | 33,373 | 3,709 | 4,121 | 44.84% |
| Ptacek | 51,009 | 5,668 | 6,298 | 49.50% |
| SemEval-18 | 3,398 | 378 | 780 | 49.12% |

Table 1 summarizes the statistics of the four datasets. The iSarcasm [5] dataset is imbalanced while SemEval-18 [9] dataset is balanced. The two source datasets are more than ten times the size of the target datasets. For all datasets, we use the predefined test set and use a random 10% split of the training set as the validation set.

## B. Preprocessing

Preprocessing is an integral part of any NLP study. It is done so that the preprocessed part would not give any weight and biasness to the experiments. There are four steps of preprocessing conducted in this study.

First of all, the datasets were preprocessed using the lexical normalization tool for tweets [26]. We then cleaned the four datasets by dropping all the duplicate tweets within and across datasets, and trimmed the texts to a maximum length of 100.

Next, we only use train and validation datasets, so we merge the train and validation datasets and use the test data as validation data. We then drop the unused column in datasets and converted all the text in datasets to lowercase.

Then, we cleaned the tweet texts by deleting all the texts inside the bracket which contain URL links, hashtags, foreign language characters, stop words removal, non-English ASCII character, and emoji. Lastly, we perform data augmentation.

Data augmentation was the crucial step in this study. We use data augmentation to increase the sarcastic labeled text to make datasets more balanced. We then observe the performance impact when applying data augmentation. To make a conclusive report, we apply data augmentation in the sarcasm labeled text to all four datasets, regardless of whether it is balanced or not, this way we can conclude the impacts. GloVe [10] was used for word embeddings to generate more data by replacing words in sentences that have similar meanings. We perform data augmentation to increase sarcasm labeled text data with 3 different sizes from 10%, 20%, and 30%.

## C. Modeling

In this paper, we experiment using advanced state-of-the-art methodologies RoBERTa [11] and using Glove [10] word embeddings for data augmentation. We use training dataset to tune the hyper-parameters of RoBERTa model. For bigger automatically-collected datasets such as Ptacek [8] and Ghosh [7], we use a batch size of 32 and an epoch size of 8, while for the smaller two manually annotated datasets include iSarcasm [5] and SemEval-18 [9] we use a batch size of 16 and an epoch size of 13. We set manual_seed parameters to 128 for all four datasets to provide reproducible results. For other parameters, we set it identically with all four datasets which is, max_seq_length: 40, learning_rate: 0.00001, weight_decay: 0.01, warmup_ratio: 0.2, max_grad_norm: 1.0, and fp16 set to True.

**TABLE 2. Hyper-parameter settings.**

| Hyper-Parameter | Value |
|---|---|
| max_seq_length | 40 |
| learning_rate | 0.00001 |
| weight_decay | 0.01 |
| warmup_ratio | 0.2 |
| max_grad_norm | 1.0 |
| num_train_epochs | 8,13 |
| train_batch_size | 16,32 |
| fp16 | true |
| manual_seed | 128 |

## IV. RESULTS AND DISCUSSIONS

In this section, the results produced by all the experiments are given. This started with the explanation of how each configuration impacting the results in this experiment and followed by explaining the confusion matrix with a true positive, true negative, false positive, and false negative.

### A. Model Performance

The performance results in this experiment are shown in terms of F-score and MCC.

**TABLE 3. F-Score performance comparison for non-augmented and augmented datasets.**

| Dataset | F - Score | | | |
|---|---|---|---|---|
| | Non-augmented | 10% augmented | 20% augmented | 30% augmented |
| iSarcasm | 0.3720 | 0.3809 | **0.4044** | 0.3828 |
| Ghosh | **0.7964** | 0.7830 | 0.7758 | 0.7835 |
| Ptacek | 0.8705 | **0.8738** | 0.8727 | 0.8717 |
| SemEval-18 | 0.6606 | 0.6666 | 0.6707 | **0.6746** |

**TABLE 4. MCC performance comparison for non-augmented and augmented datasets.**

| Dataset | MCC | | | |
|---|---|---|---|---|
| | Non-augmented | 10% augmented | 20% augmented | 30% augmented |
| iSarcasm | 0.2789 | 0.2964 | **0.3084** | 0.2939 |
| Ghosh | **0.6438** | 0.6284 | 0.6193 | 0.6294 |
| Ptacek | 0.7411 | **0.7491** | 0.7469 | 0.7442 |
| SemEval-18 | 0.4128 | 0.4286 | 0.4362 | **0.4382** |

Based on the F-score results, it can be observed that iSarcasm [5], a small and very unbalanced dataset shows a moderate performance increase by 3.2% at the 20% increase in data augmentation. SemEval-18 [9], a small and very balanced dataset shows a little performance increase by 1.4% at the 30% increase of data augmentation. On the other hand, Ptacek [8], a big and very balanced dataset shows a little performance increase by 1.4% at the 30% increase in data augmentation. Ghosh [7], a big and moderately balanced dataset shows a decrease in performance when applying data augmentation. Correspondingly, the highest performance impact for the MCC results gives similar results from the F-score performance.

Using data augmentation in a small and very unbalanced dataset like iSarcasm [5] can increase the performance of F-Score and MCC results. This happens due to the fact that by using pre-trained word embeddings such as GloVe in data augmentation, it generates new sarcastic data by replacing a word in the sentence with a similar meaning word. The newly generated data also have a similar context to the original one which prevents giving more noise to the dataset and overfitting. The augmented dataset will produces more balanced data, therefore increased the performance results, especially in detecting non-sarcasm data.

*B. Confusion Matrix Analysis*

The true positive (TP) result is noticed when the predicted tweet is found to be sarcastic, and the result of the classification shows exactly sarcastic after the experimental evaluation. True negative (TN) The true negative result is obtained when the predicted tweet is non-sarcastic, and the classification result also validates it as non-sarcastic. False positive (FP) occurs in a situation where a true negative result is obtained when the predicted tweet is non-sarcastic, but the classification result indicates that the tweet is sarcastic. False negative (FN) occurs when the true positive result is obtained when the predicted tweet is sarcastic, but the classification result shows that tweet is non-sarcastic.

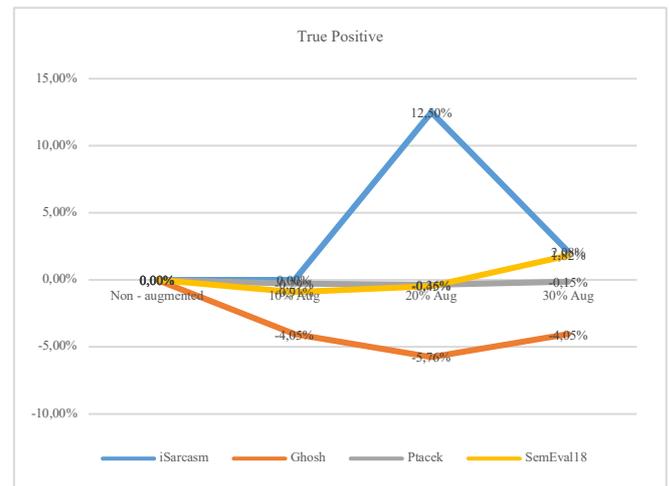

Fig. 1. True Positive results comparison.

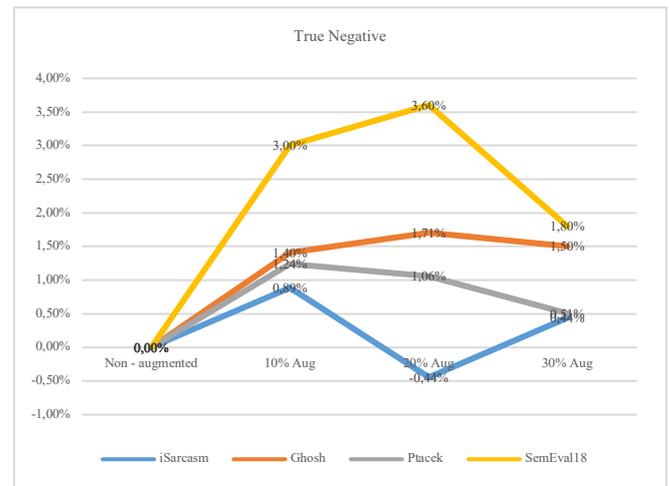

Fig. 2. True Negative results comparison.

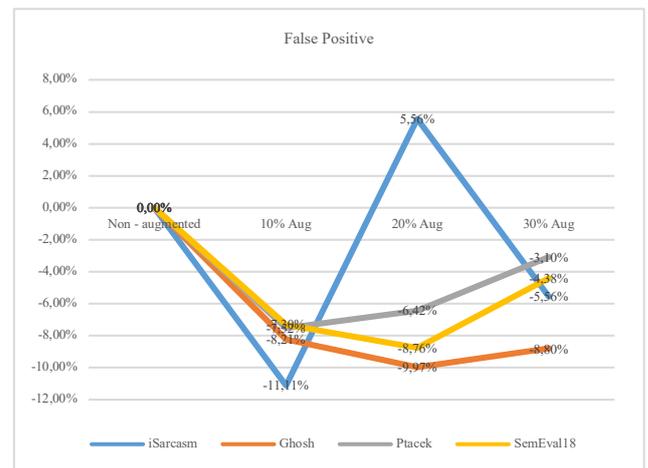

Fig. 3. False Positive results comparison.

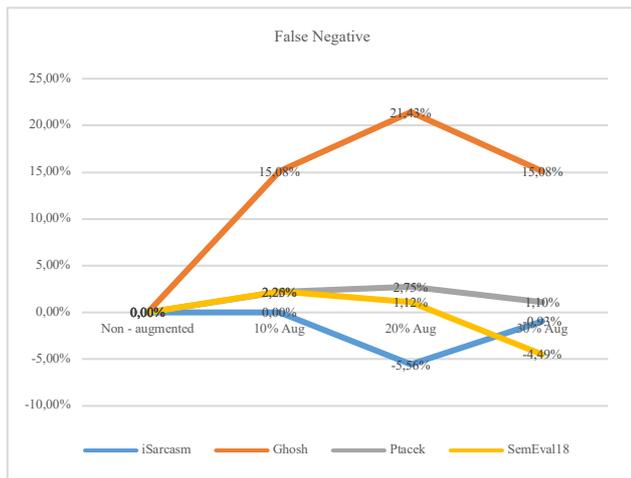

Fig. 4. False Negative results comparison.

Based on the true positive result in Fig. 1, it can be observed that in the iSarcasm [5] dataset, the amount of true positive increased mostly by 12.50% at 20% increase in data augmentation. The result means that the amount of correctly predicted as sarcastic was increased in the iSarcasm [5] dataset. However, other datasets such as Ghosh [7] and Ptacek [8] show no significant difference, while SemEval-18 [9] dataset shows a greatly reduced true positive score by -5.76% at 20% increase data augmentation.

The true negative result in Fig. 2, shows that in SemEval18 [9] dataset, there true negative increased by 3.60% at 20% increase data augmentation, meaning the amount of correctly predicted as non-sarcastic was increased. In addition, other datasets such as iSarcasm [5], Ghosh [7], and Ptacek [8] occasionally show similar trends. It means augmenting the data will helps the model learning non-sarcastic data.

The false positive result in Fig. 3, shows that overall, data augmentation reduced the amount of false positive in all datasets. This means that the model is more capable of predicting the non-sarcastic text or give a wrong result of predicting the non-sarcastic text as sarcastic. There is an interesting result shown in the iSarcasm dataset [5]. In the 20% increase data augmentation, there was a 5.56% increase in false positive results which most likely occurred because, at a 20% increase data augmentation, the iSarcasm dataset [5] had a high increase in true positive result, thus impacting the false positive result.

The false negative result in Fig. 4 shows that in the iSarcasm [5] dataset, at 20% increase data augmentation there is 5.56% of reduced false positive prediction. It means the model reduces the wrong prediction to classify sarcastic text as non-sarcastic. The same result can also be seen in SemEval-18 [9] dataset with 30% increase in data augmentation. There is a 4.49% reduced false positive prediction. However, most of these scenarios make the models increase their false negative predictions.

## V. CONCLUSION

In this research, we dealt with four datasets and augmented the data using three different data augmentation scenarios. From the experiment results, it can be concluded that in general data augmentation is not significantly impacting the model performance in predicting sarcastic text but, it increased the model capability to predict non-sarcastic text. Based on this experiment, we found that the improvement of F-score and MCC because the model is more sensitive to predict non-sarcastic data.

In addition, in a small and unbalanced dataset like iSarcasm [5], by using the right amount of data augmentation and choosing the right word embedding model, it was found there will be a significant performance increase as high as 12.50% in true positive and 3.2% F-score result.